\documentclass[sigconf, nonacm]{acmart}
\AtBeginDocument{%
  }

\setcopyright{acmlicensed}
\copyrightyear{2026}
\acmYear{2026}
\acmDOI{XXXXXXX.XXXXXXX}
\acmConference[KDD'26]{Proceedings of the 32th ACM SIGKDD Conference on Knowledge Discovery and Data Mining}{August 9-13, 2026}{Jeju, Korea}
\acmISBN{978-1-4503-XXXX-X/2018/06}
\usepackage{multirow}
\usepackage{algorithm}
\usepackage{algorithmic}

\usepackage{array}
\usepackage{textcomp}
\usepackage{stfloats}
\usepackage{url}
\usepackage{verbatim}
\usepackage{graphicx}

\usepackage{pifont}
\usepackage{float}
\usepackage{fixfoot}
\usepackage{booktabs}
\usepackage{caption}
\usepackage{subfigure}
\usepackage{enumitem}
\setlist[itemize]{leftmargin=*}
\DeclareFixedFootnote{\MLFootnote}{\url{https://grouplens.org/datasets/movielens/}}
\DeclareFixedFootnote{\lastfmFootnote}{\url{https://grouplens.org/datasets/hetrec-2011/}}
\begin{document}

%%
%% The "title" command has an optional parameter,
%% allowing the author to define a "short title" to be used in page headers.
\title{Graph Federated Learning for Personalized Privacy Recommendation}

%%
%% The "author" command and its associated commands are used to define
%% the authors and their affiliations.
%% Of note is the shared affiliation of the first two authors, and the
%% "authornote" and "authornotemark" commands
%% used to denote shared contribution to the research.
\author{Ce Na}
% \authornote{Both authors contributed equally to this research.}

% \orcid{1234-5678-9012}
\affiliation{%
  \institution{Yangzhou University}
  \city{Yangzhou}
  \state{Jiangsu}
  \country{China}
}
\email{MX120230598@stu.yzu.edu.cn}

\author{Kai Yang}
\affiliation{%
  \institution{Yangzhou University}
  \city{Yangzhou}
  \state{Jiangsu}
  \country{China}
}
\email{yangk@fudan.edu.cn}

\author{Dengzhao Fang}
\affiliation{%
  \institution{Jilin University}
  \city{Changchun}
  \state{Jilin}
  \country{China}
}
\email{hongjian318@gmail.com}

\author{Yu Li}
\affiliation{%
  \institution{Jilin University}
  \city{Changchun}
  \state{Jilin}
  \country{China}
}
\email{liyu90@jlu.edu.cn}

\author{Jingtong Gao}
\affiliation{%
  \institution{City University of Hong Kong}
  % \city{Hong Kong}
  \state{Hong Kong}
  \country{China}
}
\email{jt.g@my.cityu.edu.hk}

\author{Chengcheng Zhu}
\affiliation{%
  \institution{Nanjing University}
  \city{Nanjing}
  \state{Jiangsu}
  \country{China}
}
\email{chengchengzhu2022@126.com}

\author{Jiale Zhang}
\affiliation{%
  \institution{Yangzhou University}
  \city{Yangzhou}
  \state{Jiangsu}
  \country{China}
}
\email{jialezhang@yzu.edu.cn}

\author{Xiaobing Sun}
\affiliation{%
  \institution{Yangzhou University}
  \city{Yangzhou}
  \state{Jiangsu}
  \country{China}
}
\email{xbsun@yzu.edu.cn}

\author{Yi Chang}
\affiliation{%
  \institution{Jilin University}
  \city{Changchun}
  \state{Jilin}
  \country{China}
}
\email{yichang@jlu.edu.cn}

%%
%% By default, the full list of authors will be used in the page
%% headers. Often, this list is too long, and will overlap
%% other information printed in the page headers. This command allows
%% the author to define a more concise list
%% of authors' names for this purpose.
\renewcommand{\shortauthors}{Ce Na et al.}

\settopmatter{printacmref=false}
\renewcommand\footnotetextcopyrightpermission[1]{}
%%
%% The abstract is a short summary of the work to be presented in the
%% article.
\begin{abstract}
Federated recommendation systems (FedRecs) have gained significant attention for providing privacy-preserving recommendation services. 
However, existing FedRecs assume that all users have the same requirements for privacy protection, i.e., they do not upload any data to the server. The approaches overlook the potential to enhance the recommendation service by utilizing publicly available user data. In real-world applications, users can choose to be private or public. Private users' interaction data is not shared, while public users' interaction data can be shared. Inspired by the issue, this paper proposes a novel Graph Federated Learning for Personalized Privacy Recommendation (GFed-PP) that adapts to different privacy requirements while improving recommendation performance. GFed-PP incorporates the interaction data of public users to build a user-item interaction graph, which is then used to form a user relationship graph. A lightweight graph convolutional network (GCN) is employed to learn each user's user-specific personalized item embedding. To protect user privacy, each client learns the user embedding and the scoring function locally. Additionally, GFed-PP achieves optimization of the federated recommendation framework through the initialization of item embedding on clients and the aggregation of the user relationship graph on the server.
Experimental results demonstrate that GFed-PP significantly outperforms existing methods for five datasets, offering superior recommendation accuracy without compromising privacy. This framework provides a practical solution for accommodating varying privacy preferences in federated recommendation systems. 

% The code is available\footnote{\url{https://anonymous.4open.science/r/GFed-PP}}. 
\end{abstract}

%%
%% The code below is generated by the tool at http://dl.acm.org/ccs.cfm.
%% Please copy and paste the code instead of the example below.
%%
% \ccsdesc[500]{Do Not Use This Code~Generate the Correct Terms for Your Paper}
% \ccsdesc[300]{Do Not Use This Code~Generate the Correct Terms for Your Paper}
% \ccsdesc{Do Not Use This Code~Generate the Correct Terms for Your Paper}
% \ccsdesc[100]{Do Not Use This Code~Generate the Correct Terms for Your Paper}
% \ccsdesc[500]{Computing methodologies~Machine learning}
%%
%% Keywords. The author(s) should pick words that accurately describe
%% the work being presented. Separate the keywords with commas.
%% A "teaser" image appears between the author and affiliation
%% information and the body of the document, and typically spans the
%% page.
% \begin{teaserfigure}
%   \includegraphics[width=\textwidth]{sampleteaser}
%   \caption{Seattle Mariners at Spring Training, 2010.}
%   \Description{Enjoying the baseball game from the third-base
%   seats. Ichiro Suzuki preparing to bat.}
%   \label{fig:teaser}
% \end{teaserfigure}

% \received{20 February 2007}
% \received[revised]{12 March 2009}
% \received[accepted]{5 June 2009}

%%
%% This command processes the author and affiliation and title
%% information and builds the first part of the formatted document.
\maketitle

\section{Introduction}
Recommendation models \cite{recommender2013,recs1,recs2} play a critical role in filtering out irrelevant content and uncovering user interests. Most existing Recommendation systems \cite{lian2014geomf,liu2023linrec,gao2024smlp4rec} rely on centralized management of user data, which poses substantial privacy risks and has sparked widespread societal concern. In response, privacy regulations such as the General Data Protection Regulation (GDPR) \cite{gdpr2016general} have been introduced to enhance the protection of personal data. This highlights the growing demand for privacy-preserving recommendation systems that maintain strong performance.

Federated learning \cite{FLsurvey,FLadvances,zhang2021badfss,cz2024flpurifier,zhu2025bdpfl} is a promising approach for privacy-preserving in machine learning. It has been widely adopted in recommendation systems, leading to the development of federated recommendation systems (FedRecs) \cite{FedMF,FedNCF,PFedRec,he2024co,2024towards,zhang2025multifaceted,li2025personalized,jiang2025tutorial,zhou2025joint}. In FedRecs, recommendation models are trained locally on user devices (clients), while a central server coordinates the training process by synchronizing and aggregating model parameters. Since user data never leave the devices or become accessible to external entities, FedRecs effectively preserves user privacy.

Research on FedRecs can be divided into matrix factorization-based FedRecs (MF-FedRecs) \cite{2019federatedcf,fedreclin2020,FedMF} and graph neural network-based FedRecs (GNN-FedRecs) \cite{federatedgnn000,fedgnn111,fedgnn222,fedgnn,federatedhgnn}. MF-FedRecs primarily learn a global item embedding by collaboratively training local first-order user-item interaction matrices distributed on different devices.
However, these methods \cite{ammad2019federated,FedMF} are limited by their inability to effectively capture complex high-order structural dependencies, which are critical for improving recommendation accuracy. In contrast, GNN-based recommendation systems \cite{fedgnn,federatedhgnn} have recently achieved state-of-the-art results due to their exceptional ability to effectively capture high-order graph structural information, which offers a significant advantage over matrix factorization methods.
In the FedRecs setting, each device only has a first-order user ego graph. This graph contains only the items with which the user has directly interacted. Therefore, the main challenge for GNN-FedRecs \cite{federatedgnn000,fedgnn222} lies in learning high-order graph information while ensuring user privacy. For example, FedGNN \cite{fedgnn} proposes using a trusted third-party server to construct higher-order graphs. FedHGNN \cite{federatedhgnn} uses shared heterogeneous graphs to establish relationships between users. GPFedRec \cite{GPFedRec} constructs a user graph on the server using the similarity between items.

\begin{figure*}[!t]
\centering
\includegraphics[width=\textwidth]{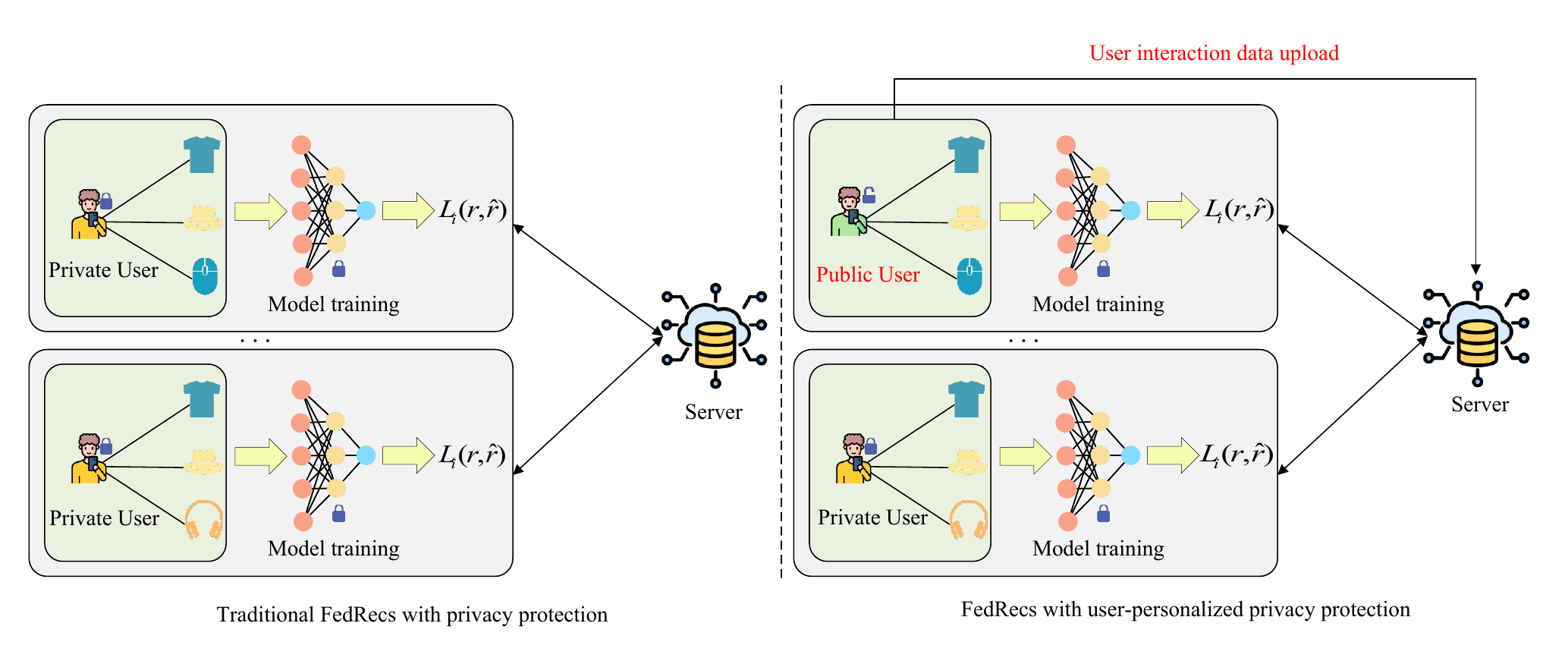}
\caption{Traditional FedRecs with privacy protection v.s. FedRecs with user-personalized privacy.
}
\label{fig_2}
\end{figure*}

Existing FedRecs architectures \cite{federated2023,federatedzhang2023comprehensive} typically assume uniform privacy preferences across all users, overlooking the fact that users may have different privacy requirements. In real-world platforms such as Douyin, users can choose to be public or private. Public users’ interaction data can be shared and utilized, while private users’ data remains local. However, current FedRec methods \cite{FedMF,FedNCF} fail to take advantage of publicly available interaction data, missing opportunities to improve recommendation quality. Enhancing user experience is a core objective of recommendation services. Enforcing a uniform privacy policy that prohibits all users from uploading interaction data may lead to degraded model performance, thereby reducing user satisfaction and platform revenue. To address this, it is crucial to design a federated recommendation framework that selectively leverages interaction data from public users while respecting the privacy of others. Figure \ref{fig_2} illustrates the comparison between conventional federated privacy protection recommendation approaches and a federated personalized privacy protection framework. Unlike traditional methods, the proposed approach incorporates publicly available user interactions into the training process, enabling more accurate recommendations without compromising sensitive information. This paper explores how to effectively utilize public interaction data to enhance recommendation performance while maintaining user-level privacy control.

In this paper, we propose Graph Federated Learning for Personalized Privacy Recommendation (GFed-PP), a unified federated recommendation framework designed to accommodate users with varying privacy preferences, including both public and private users. GFed-PP utilizes publicly available user-item interaction data to construct a server-side user relationship graph, enabling the modeling of user similarities while ensuring that private data remains unexposed. To generate user-personalized item embeddings, a lightweight graph aggregation mechanism is used to integrate information from behaviorally similar public users. To ensure privacy protection, user embeddings and scoring functions are retained locally on each client, with only item embeddings being communicated to the server. The framework further differentiates between public and private users by assigning personalized item embeddings to public users, while delivering globally shared item embeddings to private users. These design strategies allow GFed-PP to effectively balance recommendation accuracy and privacy protection, making it a practical solution for real-world federated recommendation settings with diverse user privacy demands.
In conclusion, we highlight the key contributions of this paper as follows.

\begin{itemize}
\item 
We propose for the first time a federated recommendation method that leverages public interaction data to enhance recommendation performance while addressing the distinct privacy requirements of public and private users.

\item 
We introduce a user relationship modeling strategy to capture user similarity for personalized recommendation, integrated within the unified GFed-PP framework.

\item 
We design a novel server-side item embedding distribution mechanism that delivers user-personalized item embeddings to public users, while providing global item embeddings to private users for local training.

\item 
Extensive experiments on five recommendation datasets demonstrate that GFed-PP significantly outperforms existing federated recommendation baselines.

\end{itemize}

\section{Preliminary}
In this section, we provide the necessary background to understand our proposed framework. We begin by introducing the fundamental concepts of federated recommendation systems. We then formally define the problem formulation.
\subsection{Federated Recommendation}
Let $U$ and $I$ represent the sets of users and items, respectively. The user-item interaction data are denoted as $r_{um}$,  where $u\in{U}$ and $m\in{I}$. Other notations can be found in \textbf{Appendix} \ref{aaa}. For a recommendation model $F$ parameterized by $\theta$, the prediction for user $u$ and item $m$ is given by $\hat{{r}}_{um}={F}(u,m|\theta)$. We define the user relationship graph with $G({U},{E})$, where $U$ is the set of users and $E$ is the set of edges. The adjacency matrix of $G$ is denoted as $\mathbf{A}\in\{0,b\}^{N\times N}$, where $N$ is the number of users in $U$. Specifically, $\mathbf{A}{uv} = b$ indicates an edge between users $u, v \in \mathcal{U}$, where $b$ denotes the number of items both users have interacted with. If no such overlap exists, $\mathbf{A}{uv} = 0$.

For a federated recommendation model $F$ parameterized by $\theta$, the goal is to predict user $u$'s preference on item $m$ as $\hat{{r}}_{um}={F}(u,m|\theta^*)$, and the optimal model parameter $\theta^*$ are learned by minimizing the accumulated loss across all local models. Specifically, $\theta^*$ = argmin$_{\theta} \sum_{i=1}^{N} \omega_i {L}_i(\theta)$,  where ${L}_i(\theta)$ is the loss function of the $i$-th client, and $\omega_i$ is the corresponding client weight. Additionally, $I_i$ and $I_i^-$ represent the set of interacted positive items and the set of sampled negative items for $i$-th client.

\subsection{Problem Formulation}

\subsubsection{Federated Learning Optimization Objective}
We treat each user as a client in the federated learning framework. The recommendation task can be described as a personalized federated learning problem, aiming to provide personalized services for each user.
We utilize a neural network recommendation model ${M}_\theta$ consisting of three components, the user embedding module parameterized by ${p}$, the item embedding module parameterized by ${q}$, and the score function module parameterized by ${o}$. The score function module predicts a user's rating based on the embeddings of both the user and the item.
In particular, since item embeddings do not contain privacy-sensitive information, they can transfer shared knowledge between users. On the other hand, user embeddings and the score function contain sensitive user data and are maintained locally to capture user personalization. We formulate the proposed GFed-PP as the below optimization objective,
\begin{equation}
\min_{\{\theta_1,...,\theta_N\}}\quad\sum_{i=1}^N L_i(\theta_i),
\label{eq: optimization}
\end{equation}
where $\theta_{i} = {p_{i}, q_{i}, o_{i}}$ denotes the model parameters of the $i$-th client, comprising three components: (1) $p_i$, the user embedding module parameters; (2) $q_i$, the item embedding module parameters; and (3) $o_i$, the parameters of the score function module.
\subsubsection{Recommendation Model Loss Function}
In this paper, we focus on a common recommendation scenario that relies solely on implicit user-item interaction data. Specifically, we define ${r}_{um}=1$ if user ${u}$ has interacted with item ${m}$, and ${r}_{um}=0$ otherwise. The original feature representations of users and items are assumed to be unavailable. Given the binary nature of the implicit feedback, we adopt the binary cross-entropy loss to optimize the recommendation model for the $i$-th client, formulated as:
\begin{equation}{L}_i(\theta_i;{r}_{um},\hat{{r}}_{um})=-\sum_{(u,m)\in I_i}\log\hat{{r}}_{um}-\sum_{(u,m^{\prime})\in I_i^-}\log(1-\hat{{r}}_{um^{\prime}}),
\label{eq:2}
\end{equation}
where $I_i$ and $I_i^-$ denote the sets of positive (interacted) items and sampled negative items for the $i$-th client, respectively. The predicted rating $\hat{r}_{um}$ is the output of the model for user $u$ and item $m$. To construct $I_i^-$ efficiently, negative samples are drawn from the set of unobserved user-item interactions, according to a predefined negative sampling ratio.

\section{Method}
\begin{figure*}[!t]
\centering
\setlength{\fboxsep}{0pt} % Remove the frame around the image
\includegraphics[width=7in]{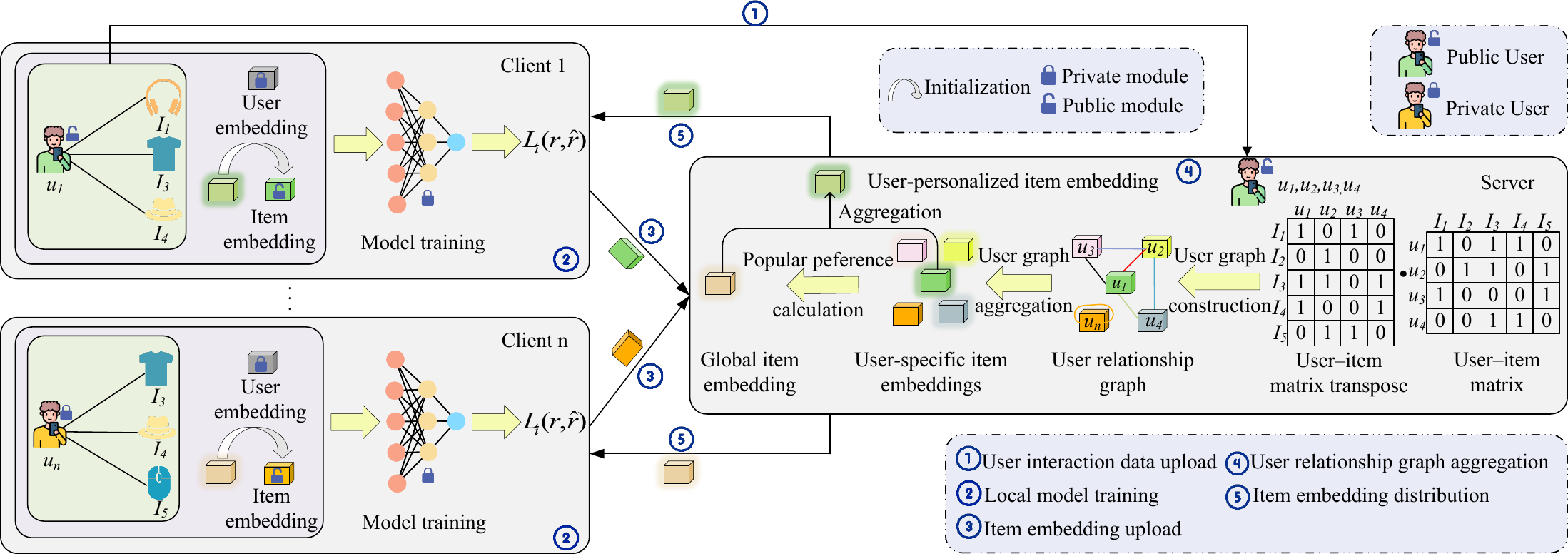}
\caption{The framework of GFed-PP. There are five steps in each communication round: \ding{192} public users upload interaction graph. \ding{193} Clients train local recommendation models for public and private users, initializing item embeddings with user-personalized and global embeddings, respectively. \ding{194} Clients upload updated item embeddings to the server. \ding{195} The server constructs a user relationship graph from public user interactions, derives user-specific item embeddings, computes global item embeddings capturing popular preferences, and aggregates them for personalized embeddings. \ding{196} The server distributes global item embedding to private users and user-personalized item embedding to public users for the next round of optimization.
}
\label{fig_1}
\end{figure*}

In this section, we introduce the Graph Federated Learning framework for Personalized Privacy Recommendation (GFed-PP). As illustrated in Figure~\ref{fig_1}, users are categorized into two groups: public users, who share their interaction data, and private users, who retain their data locally. The framework operates as follows:
\ding{192} Public users upload their interaction graphs to the server, whereas private users keep their interaction graphs local to preserve privacy. 
\ding{193} Each client trains a local recommendation model for both public and private users. For private users, local item embeddings are initialized using the global item embedding, capturing popular preferences. For public users, local item embeddings are initialized with user-personalized item embeddings, tailored to individual preferences. The local model is optimized using the recommendation task loss $\mathcal{L}_i(r, \hat{r})$ as the objective.
\ding{194} In each training round, clients upload their locally trained item embeddings to the server after completing local recommendation model training.
\ding{195} The server constructs a user relationship graph from the interaction graphs of public users and employs a lightweight Graph Convolutional Network (GCN)~\cite{gcn} to aggregate neighbor embeddings, generating user-specific item embeddings. Additionally, a global item embedding is computed to reflect popular preferences. By leveraging the user relationship graph, the server integrates neighbor preferences with global preferences to produce personalized item embeddings for each user.
\ding{196} Finally, user-personalized item embeddings are distributed to public users, while the global item embedding is sent to private users, enabling personalized local model training.
This framework effectively balances recommendation accuracy and user privacy by leveraging public user data and global preferences while ensuring private user data remains local.

In the following, we present our proposed GFed-PP framework in detail. We begin with the overall training procedure that coordinates both client-side and server-side updates across public and private users. Then, we describe how the server leverages shared public interaction data to construct a user relationship graph and generate user-personalized item embeddings via lightweight graph convolution. Subsequently, we introduce the personalized client-side learning strategy, which incorporates these embeddings into personalized model updates. Finally, we enhance the privacy protection of GFed-PP by integrating local differential privacy mechanisms \cite{ldp} into the training pipeline.
\subsection{Framework of GFed-PP}
This section presents the overall architecture of GFed-PP, a novel federated recommendation framework designed to enhance both personalization and privacy. GFed-PP aims to address two key challenges in federated recommendation: (1) how to enrich item representations without exposing sensitive user data, and (2) how to capture both individual and global preferences under privacy constraints. To this end, GFed-PP alternates between client-side personalization and server-side graph-based aggregation.

Specifically, two alternate steps are performed to achieve the optimization objective described in Eq. \eqref{eq: optimization}. \textbf{\textit{First}}, for the clients, public users upload their historical interaction data along with the post-trained item embeddings, while private users only upload the post-trained item embeddings. \textbf{\textit{Second}}, for the server, it constructs a user relationship graph based on interaction data received from public users, then performs graph aggregation to obtain item embeddings for specific users. The global item embedding is obtained by averaging user-specific item embeddings to represent popular preferences. Finally,
the server aggregates the global item embedding with the item embeddings of public users to generate user-personalized item embeddings for public users. Private users cannot participate in user graph aggregation, so the server distributes the global embedding to private users.
% The optimization procedure is illustrated in Algorithm \ref{alg: GFed-PP} for \textbf{Appendix} \ref{alg}. 

\subsection{Server-Side Update via User Relationship Graph}
The objective of this module is to enhance item embeddings by leveraging collaborative knowledge distilled from public users, while strictly preserving the privacy of private users. This design addresses the challenge of limited data availability in federated settings by introducing a user relationship graph to enable richer and more informative representation learning.

To construct a user graph, the server receives the interaction graph uploaded by public users and builds the corresponding user-item interaction matrix. By multiplying this matrix with its transpose, a user relationship graph $G$ iis derived, whose adjacency matrix captures the similarity between public users based on their co-interaction items. Formally, the adjacency matrix is defined as: 
\begin{equation}
\mathbf{A}_{uu} = \mathbf{A}_{um} {\mathbf{A}_{um}}^T,
\label{eq: user graph}
\end{equation}
where the entry \(\mathbf{A}_{uu}[i,j]\) represents the sum of the items of the interaction values of users \(i\) and \(j\) across all items, effectively measuring their co-interaction frequency. Consequently, users with more shared co-interaction items receive higher similarity weights, reflecting their co-engagement across items.

The $\mathbf{A}_{uu}$ obtained from the above formula contains self-loops, for public users, we remove the self-loops to eliminate the relationships between users and themselves, thereby preventing excessive weight on self-information. For privacy users, we add the self-loops to ensure self-information is reserved. The matrix $\mathbf{A}$ of the user relationship graph for all users is derived and then normalized as,
\begin{equation}
    \mathbf{\tilde{A}}={\mathbf{D}^{-\frac12}\mathbf{A}\mathbf{D}^{-\frac12}},
\label{eq: nor}
\end{equation}
where $\mathbf{D}$ is degree matrix of $\mathbf{A}$ and $\mathbf{D}_{ii}=\sum_{j}\mathbf{A}_{ij}$. 

As user features contain sensitive information and cannot be uploaded to the server, we use the item embeddings uploaded by each user as their features. The lightweight GCN is applied to obtain user-specific item embeddings for all users. The convolution operation is shown as follows, 
\begin{equation}
    \mathbf{S}=\mathbf{\tilde{A}}^l\mathbf{Q},
\label{eq: gcn}
\end{equation}
where $\mathbf{Q}$ represents the initial item embedding matrix whose $i$-th row corresponds to the item embedding $q_i$ provided by user $i$, and $\mathbf{S}$ denotes the aggregated item embeddings(user-specific item embeddings), with the $i$-th row being $s_i$. Furthermore, $l$ refers to the number of convolution layers. We set $l$ = 1 in this paper. 

In lightweight GCN aggregation, users with more neighbors participate in more frequent calculations of $s_i$. To capture the popular preference, we calculate the global item embedding by averaging all $s_i$, The calculation is formulated as follows,
\begin{equation}
    q_g=\frac{1}{n}\sum_{i=1}^{n}s_{i},
\label{eq: global}
\end{equation}
where $s_i$ represents the item embedding of the $i$-th row user, and $n$ is the total number of users. The averages we take on item embeddings after user graph aggregation, which focuses more on popular preferences. This approach achieves better performance relative to the absence of user graph aggregation.

Public users can enrich their item embeddings by accessing the item information of other public users through the user relationship graph. We aggregate the user-specific item embedding with the global item embedding to get the user-personalized item embeddings $Q_u$, which are then provided to public users. This ensures a balance between personal preferences and popular preferences. Private users cannot update their item embeddings through the user relationship graph, we provide private users with global item embeddings $q_g$ to capture global preferences. Therefore, the user-personalized item embedding for each public user distributed by the server is as follows,
\begin{equation}
    q_{u_i}=\alpha s_{i}+(1-\alpha)q_g,
\label{eq: personal}
\end{equation}
where $\alpha \in [0,1]$ denotes the degree of personalization. Since the proportion of public users may vary across scenarios, different values of $\alpha$ can be adapted accordingly.
\subsection{Personalized Local Update on Clients}
This module personalizes the model for each client by fine-tuning local parameters with the item embeddings provided by the server. It ensures user-level adaptation while maintaining privacy.

Specifically, each client receives global item embeddings or user-personalized item embeddings from the server, which are then used to initialize the local item embedding parameters for subsequent fine-tuning. Public users receive user-personalized item embeddings $\mathbf{Q}_u$ depicting personalized preferences, while private users receive global item embedding $q_g$ depicting popular preferences. The $\mathbf{Q}_u$ and $q_g$ are integrated into the local model training process. Specifically, we first initialize item embedding $q_i$ for every private user using the global item embedding $q_g$, and initialize the item embedding $q_i$ for every public user using user-personalized item embedding $q_{u_i}$. Both the user embedding $p_i$ and the score function $o_i$ are inherited from the trained model in the last round for all clients. Then, we train the local recommendation model by Eq. \eqref{eq:2}.

We update the $\theta_{i}$ using the stochastic gradient descent algorithm, with the $t$-th update step expressed as follows,
\begin{equation}
    \theta_i^{t+1}=\theta_i^t-\eta\partial_{\theta_i^t}{L}_{i},
\label{eq: sgd}
\end{equation}
where $\eta$ represents the learning rate and $\partial_{\theta_i^t}{L}_{i}$ denotes the gradient of the loss for the model parameters.

\subsection{Privacy Protection Enhanced GFed-PP}
This component enhances the privacy guarantees of GFed-PP by introducing formal differential privacy mechanisms. It addresses potential leakage risks from shared item embeddings.

Although GFed-PP adheres to the federated learning paradigm by retaining all user private data on local devices, thereby substantially reducing the risk of direct privacy breaches, this alone is not sufficient. Since item embeddings are still transmitted to the server during training, additional safeguards are necessary to prevent potential indirect leakage.  To further address privacy concerns associated with the transmission of item embeddings, we incorporate a local differential privacy (LDP) mechanism \cite{ldp} to perturb the embeddings before they are uploaded to the server. Specifically, we inject zero-mean Laplace noise into the item embeddings before it is uploaded to the server,
\begin{equation}
    q_i=q_i+Laplacian(0,\delta),
\end{equation}
where $\delta$ denotes the noise strength. It becomes more difficult to retrieve the updated items by observing the item embeddings, and privacy-preserving improves as $\delta$ increases.

In summary, our GFed-PP framework achieves privacy-preserving federated recommendation by: (1) leveraging user-uploaded interaction data to construct a global item graph on the server, enabling high-order structural modeling; (2) distilling knowledge from server-side item embeddings to locally enhance user representations; and (3) applying local differential privacy mechanisms to ensure rigorous privacy protection. The detailed training procedure is outlined in Algorithm \ref{alg: GFed-PP} for \textbf{Appendix} \ref{alg}. 

\section{Experiments}
In the section, we present experiments designed to evaluate the proposed method, intending to address the following questions,
\begin{itemize} 
\item \textbf{\textit{Q1}}: Does GFed-PP outperform existing state-of-the-art federated and centralized recommendation models? \item \textbf{\textit{Q2}}: How does each component of GFed-PP contribute to its overall performance?
\item \textbf{\textit{Q3}}: What is the impact of key hyper-parameters on the performance of GFed-PP? 
\item \textbf{\textit{Q4}}: How does the recommendation performance differ between public and private users in the GFed-PP model? 
\item \textbf{\textit{Q5}}: How robust is GFed-PP when incorporating local differential privacy techniques?
\end{itemize}
\subsection{Experimental Setup}
We evaluate the proposed GFed-PP on five widely used datasets: MovieLens-100K\MLFootnote{}, MovieLens-1M\MLFootnote{} \cite{movielens}, Lastfm-2K\lastfmFootnote{} \cite{hetrec2011}, HetRec2011\lastfmFootnote{} \cite{hetrec2011}, and Amazon-Video\footnote{\url{https://jmcauley.ucsd.edu/data/amazon/}} \cite{amazon}. MovieLens datasets provide movie ratings, with each user rating at least 20 movies. Lastfm-2K contains user–artist interactions and associated tags for music recommendation. HetRec2011 extends MovieLens-10M by linking movies to IMDb and Rotten Tomatoes. Amazon-Video includes product reviews and metadata. We filter out users with fewer than 5 interactions in Lastfm-2K and Amazon-Video.
\begin{table}[h!]
\centering
\setlength{\tabcolsep}{3pt} % 调整表格列间距
\caption{Statistics of the datasets.}
\label{table: datasets}
\begin{tabular}{lcccc}
\hline
\textbf{Dataset} & \textbf{\# Users} & \textbf{\# Items} & \textbf{\# Interactions} & \textbf{Sparsity} \\ \hline
MovieLens-100K & 943  & 1682  & 100,000   & 93.70\%   \\ 
MovieLens-1M   & 6,040 & 3,706  & 1,000,209 & 95.53\%   \\ 
Lastfm-2K      & 1,600 & 12,454 & 185,650   & 99.07\%   \\ 
HetRec2011     & 2,113 & 10,109 & 855,598   & 95.99\%   \\ 
Amazon-Video   & 8,072 & 11,830 & 63,836   & 99.93\%   \\ \hline
\end{tabular}
\end{table}

Table~\ref{table: datasets} summarizes the dataset statistics. For fair comparison, we adopt the widely used leave-one-out evaluation protocol \cite{ncf2017}, where the most recent interaction is held out for testing and the rest for training. Additionally, we reserve the second-to-last interaction as a validation sample for tuning. To reduce evaluation cost, we follow the standard practice \cite{koren2008factorization,ncf2017} of ranking the ground-truth item among 99 randomly sampled negative items. Performance is measured by Hit Ratio (HR) and Normalized Discounted Cumulative Gain (NDCG) \cite{HRNDCG}, with top-$K$ ($K=10$ by default). HR checks if the test item appears in the top-$K$ list, while NDCG assigns higher scores to higher-ranked relevant items.

\subsection{Baselines and Implementation Details}
\textbf{Baselines.} We compare our proposed federated recommendation method against a series of representative baselines under both centralized and federated settings. All compared methods operate solely on user-item interaction data without leveraging any auxiliary side information, representing the most fundamental recommendation setting. The baselines include MF \cite{MF}, NCF \cite{ncf2017}, FedMF \cite{FedMF}, FedNCF \cite{FedNCF}, FedRecon \cite{FedRecon}, MetaMF \cite{MetaMF}, FedLightGCN \cite{2020lightgcn}, FedPerGNN \cite{fedgnn}, PFedRec \cite{PFedRec}, and GPFedRec \cite{GPFedRec}. Detailed descriptions of these baselines are provided in \textbf{Appendix}~\ref{b1}.

\textbf{Implementation Details.} 
These methods are implemented using the PyTorch framework. Hyper-parameter configurations and more implementation details are in \textbf{Appendix} \ref{c1}.

\subsection{Performance Analysis \textbf{\textit{(Q1)}}}
\begin{table*}[htbp]
    \centering

    \caption{Performance comparison across five datasets. The best results are highlighted in bold, and the second-best results are underlined. "CenRec" and "FedRec" refer to centralized and federated settings, respectively. FedPerGNN could not be run on HetRec2011 and Amazon-Video due to excessive memory allocation, denoted as "–". We conduct significance test (i.e., two-sided t-test with $p$ < 0.05), and “Improvement” indicate the performance improvement over the best baseline.}
    \scalebox{0.9}{
    \begin{tabular}{clcccccccccc}
        \toprule
        & \multirow{2}{*}{Method} & \multicolumn{2}{c}{MovieLens-100K} & \multicolumn{2}{c}{MovieLens-1M} & \multicolumn{2}{c}{Lastfm-2K} & \multicolumn{2}{c}{HetRec2011} & \multicolumn{2}{c}{Amazon-Video} \\
             &  & HR@10 & NDCG@10 & HR@10 & NDCG@10 & HR@10 & NDCG@10 & HR@10 & NDCG@10 & HR@10 & NDCG@10 \\
        \midrule
       \multirow{2}{*}{CenRec} & MF       & 64.42 & 38.59 & 68.84 & 41.61 & 83.21 & 71.93 & 66.51 & 41.33 & 46.99 & 30.15 \\
        & NCF   & 64.79 & 38.11 & 64.58 & 38.14 & 82.33 & 67.46 & 65.13 & 40.15 & \underline{60.35} & \underline{39.04} \\
        \midrule
        \multirow{8}{*}{FedRec}& FedMF    & 64.97 & 39.13 & 67.88 & 40.81 & 81.29 & 69.27 & 64.89 & 40.35 & 59.54 & 38.43 \\
        & FedNCF   & 60.77 & 34.05 & 60.55 & 34.17 & 81.66 & 61.43 & 60.98 & 36.39 & 57.73 & 36.80 \\
        & FedRecon & 64.77 & 38.10 & 63.34 & 38.72 & 82.21 & 67.66 & 61.78 & 34.28 & 58.59 & 37.92 \\
        & MetaMF   & 66.52 & 40.88 & 45.81 & 25.33 & 80.77 & 63.82 & 54.11 & 31.93 & 57.76 & 37.44 \\
        & FedLightGCN & 23.58 & 12.44 & 36.43 & 14.68 & 45.77 & 16.43 & 22.32 & 7.46 & 55.76 &  34.52\\
        & FedPerGNN & 11.54 & 5.06 & 9.24 & 4.02 & 11.82 & 4.63 & $-$ & $-$ & $-$ & $-$ \\
        & PFedRec  & 71.83 & \underline{43.75} & \underline{73.24} & \underline{44.22} & 82.06 & 72.46 & 69.70 & 43.10 & 59.24 & 37.60 \\
        & GPFedRec & \underline{72.32} & 43.19 & 72.43 & 43.76 & \underline{83.15} & \underline{73.96} & \underline{70.21} & \underline{43.25} & 58.34 & 38.27 \\
        \midrule
        Ours & GFed-PP  & \textbf{73.62} & \textbf{44.06} & \textbf{73.53} & \textbf{44.41} & \textbf{90.13} & \textbf{82.79} & \textbf{73.28} & \textbf{46.37} & \textbf{82.83} & \textbf{73.61}  \\
        \midrule
        \multicolumn{2}{c}{Improvement} & $\uparrow$ 1.80\%  & $\uparrow$ 0.71\%  & $\uparrow$ 0.40\%  & $\uparrow$ 0.43\%  & $\uparrow$ 8.39\%  & $\uparrow$ 11.94\%  & $\uparrow$ 4.37\%  & $\uparrow$ 7.21\%  & $\uparrow$ 37.25\%  & $\uparrow$ 88.55\% \\
        
        \bottomrule
    \end{tabular}}
    \label{table：Performance}
\end{table*}
To evaluate the performance of our proposed GFed-PP framework on recommendation tasks, Table~\ref{table：Performance} reports the results of all compared methods on five datasets, using HR@10 and NDCG@10 as evaluation metrics. Below, we summarize the key findings and discuss their implications.

\begin{itemize}
\item \textbf{GFed-PP vs. Centralized Methods.}
GFed-PP consistently outperforms centralized recommendation methods across all five datasets. On the Amazon-Video dataset in particular, it achieves significant improvements of \textbf{37.25\%} in HR@10 and \textbf{88.55\%} in NDCG@10. Centralized models typically employ uniform item embeddings and shared scoring functions, relying solely on user embeddings to model personalization. In contrast, GFed-PP treats user embeddings and scoring functions as private, enabling client-specific representation learning. Furthermore, item embeddings are dynamically updated on the server to capture both global popularity trends and user-specific preferences, resulting in enhanced recommendation performance.

\item \textbf{GFed-PP vs. Federated Recommendation Baselines.}
GFed-PP also surpasses all federated recommendation (FedRec) baselines on every dataset. The most substantial improvements appear again on Amazon-Video, with HR@10 and NDCG@10 increasing by \textbf{39.12\%} and \textbf{91.54\%}, respectively. Most existing FedRecs share uniform item embeddings among clients, which fails to capture diverse user interests. PFedRec, the strongest baseline, achieves the second-best performance on some datasets by locally fine-tuning item embeddings. This aligns with our approach of replacing static embeddings with personalized ones. GPFedRec similarly achieves strong results by constructing user relationships based on item similarity and propagating preferences from similar users.

\item \textbf{GFed-PP vs. Federated GNN-based Methods.}
GFed-PP significantly outperforms two GNN-based FedRecs: FedLightGCN and FedPerGNN. In FedLightGCN, each client applies a GCN to its local subgraph. However, items that share the same interacting users form indistinguishable neighborhoods, resulting in homogeneous representations and degraded accuracy. FedPerGNN performs poorly in implicit feedback scenarios, especially when negative samples are introduced. Since it depends on user interactions to identify higher-order neighbors, the presence of false negatives causes performance drops, which become more severe in leave-one-out settings where more negative samples are introduced due to increased training data.
\end{itemize}

\subsection{Ablation Study \textbf{\textit{(Q2)}}}
\begin{table*}[htbp]
    \centering

    \setlength{\tabcolsep}{5pt} % 调整表格列间距
    \caption{Results of the ablation study. The notations $w/o$ IEI, $w/o$ UGC, and $w/o$ U-PIE indicate the removal of the Client Item Embedding Initialization, User Graph Construction, and User-Personalized Item Embedding components, respectively.}
    \begin{tabular}{lcccccccccc}
        \toprule
        \multirow{2}{*}{Method} & \multicolumn{2}{c}{MovieLens-100K} & \multicolumn{2}{c}{MovieLens-1M} & \multicolumn{2}{c}{Lastfm-2K} & \multicolumn{2}{c}{HetRec2011} & \multicolumn{2}{c}{Amazon-Video} \\
        & HR@10 & NDCG@10 & HR@10 & NDCG@10 & HR@10 & NDCG@10 & HR@10 & NDCG@10 & HR@10 & NDCG@10 \\
        \midrule
        ${w/o}$ IEI & 69.46 & 39.47 & 66.71 & 38.48 & 80.94 & 70.84 & 70.29 & 43.17 & 60.57 & 39.65 \\
        ${w/o}$ UGC & 70.86 & 41.91 & 70.96 & 39.34 & 82.79 & 73.62 & 72.39 & 45.55 & 59.72 & 38.58 \\
        ${w/o}$ U-PIE & 71.79 & 42.92 & 73.06 & 44.20 & 87.34 & 79.85 & 72.69 & 46.33 & 82.21 & 72.59 \\
        
        GFed-PP  & \textbf{73.62} & \textbf{44.06} & \textbf{73.53} & \textbf{44.41} & \textbf{90.13} & \textbf{82.79} & \textbf{73.28} & \textbf{46.37} & \textbf{82.83} & \textbf{73.61}  \\
        \bottomrule
    \end{tabular}
    \label{ablation study}
\end{table*}
\textit{\textbf{Impact of model components.}} 
To elucidate the contributions of the key components in our proposed GFed-PP framework, we decompose it into three primary design elements: Client Item Embedding Initialization (IEI), User Graph Construction (UGC), and User-Personalized Item Embedding (U-PIE). We conduct an ablation study by systematically removing each component, denoted as $w/o$ IEI, $w/o$ UGC, and $w/o$ U-PIE, respectively, and evaluate their impact on federated recommendation performance across five diverse datasets. The results are presented in Table~\ref{ablation study}.

The findings from the ablation study yield the following insights:
\begin{itemize} 
\item \textbf{Impact of IEI Removal}: The absence of the IEI component impairs the model's ability to capture popular user preferences, resulting in a consistent performance decline across all datasets. This underscores IEI's role in initializing item embeddings that reflect broadly shared preferences.
\item \textbf{Impact of UGC Removal}: The exclusion of UGC significantly degrades performance on the Amazon-Video dataset, while yielding only marginal performance impacts on the other four datasets. This inconsistency underscores UGC's critical role in effectively modeling rich user-item interaction patterns, particularly in datasets exhibiting dense connectivity structures.
\item \textbf{Impact of U-PIE Removal}: The removal of the U-PIE component leads to uniform performance degradation across all five datasets, emphasizing its essential role in generating personalized item embeddings that improve the relevance and quality of user-specific recommendations.
\item \textbf{Validation of GFed-PP Design}: The ablation study confirms the efficacy of GFed-PP's architecture, demonstrating that the synergistic integration of IEI, UGC, and U-PIE is crucial for effectively capturing both global popular preference trends and user-specific preferences in federated recommendation systems.
\end{itemize}

\subsection{Hyper-parameters Analysis \textbf{\textit{(Q3)}} \label{par}}
In this section, we investigate the impact of four vital hyper-parameters related to our proposed method on model performance.

\begin{figure}
	\centering
        
	\subfigure[Performance of HR] {
		\begin{minipage}[t]{0.48\linewidth}
			\centering
                
			\includegraphics[width=1\linewidth]{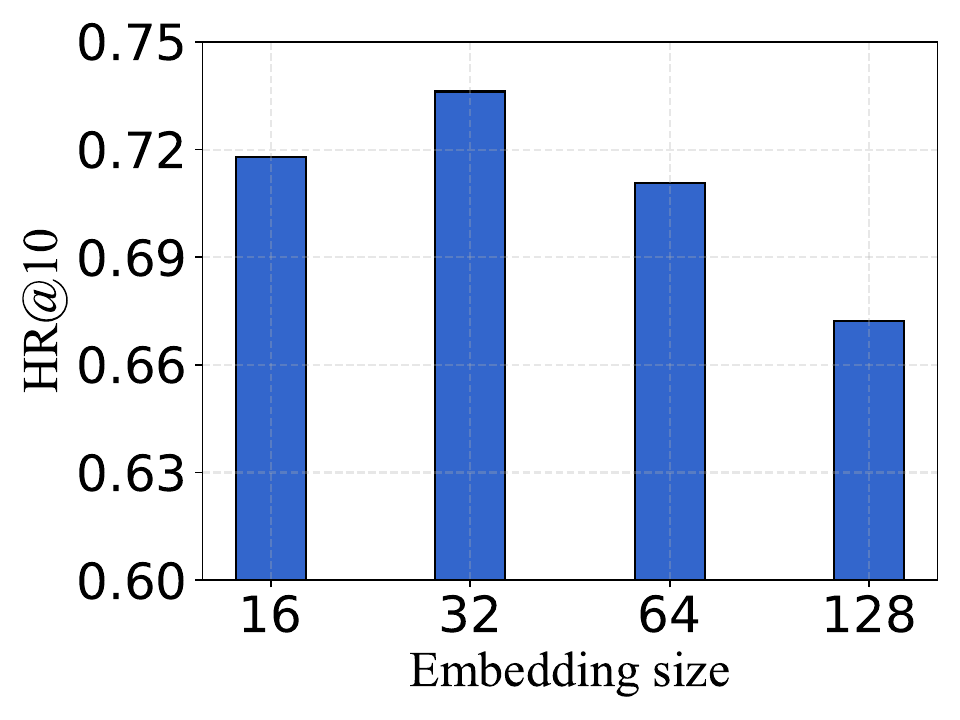}
			% \label{F5a}
	\end{minipage}}
	\subfigure[Performance of NDCG]{
		\begin{minipage}[t]{0.48\linewidth}
                \centering
                
			\includegraphics[width=1\linewidth]{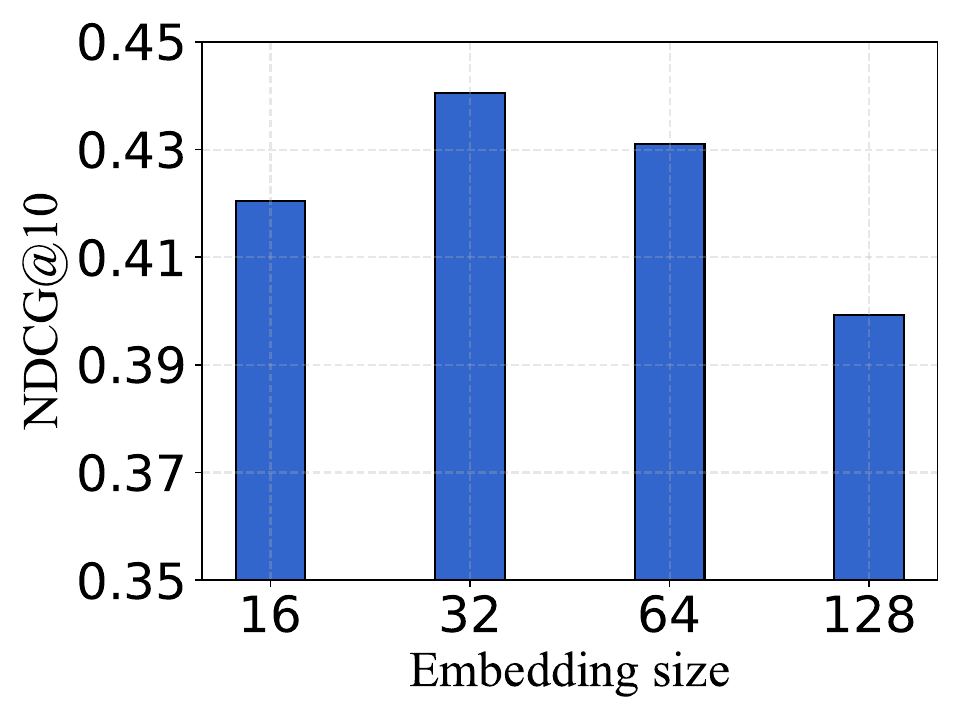}
			% \label{F5b}
	\end{minipage}}%
	\caption{Effect of embedding size. We present the HR and NDCG results for both recommendation metrics on the MovieLens-100K dataset.}
	\label{embedding}
	% \vspace{-2mm}
\end{figure}

\textit{\textbf{Size of embedding.}}
Our method utilizes a multi-layer perceptron (MLP) scoring function that integrates user and item embeddings to predict preferences. To investigate the impact of embedding dimension, we maintain a fixed MLP architecture while evaluating dimensions \( d \in \{16, 32, 64, 128\} \) as empirically validated in Figure \ref{embedding}. The model exhibits suboptimal performance at \( d = 16 \) due to constrained representational capacity. Performance peaks at \( d = 32 \) on both evaluation metrics, confirming this dimension as optimal. Further dimension increases yield progressive performance degradation: \( d = 64 \) shows measurable decline compared to the peak, while \( d = 128 \) manifests the most significant performance deterioration caused by severe overfitting.

\begin{figure}
	\centering
	\subfigure[Performance of HR] {
		\begin{minipage}[t]{0.48\linewidth}
			\centering
			\includegraphics[width=1\linewidth]{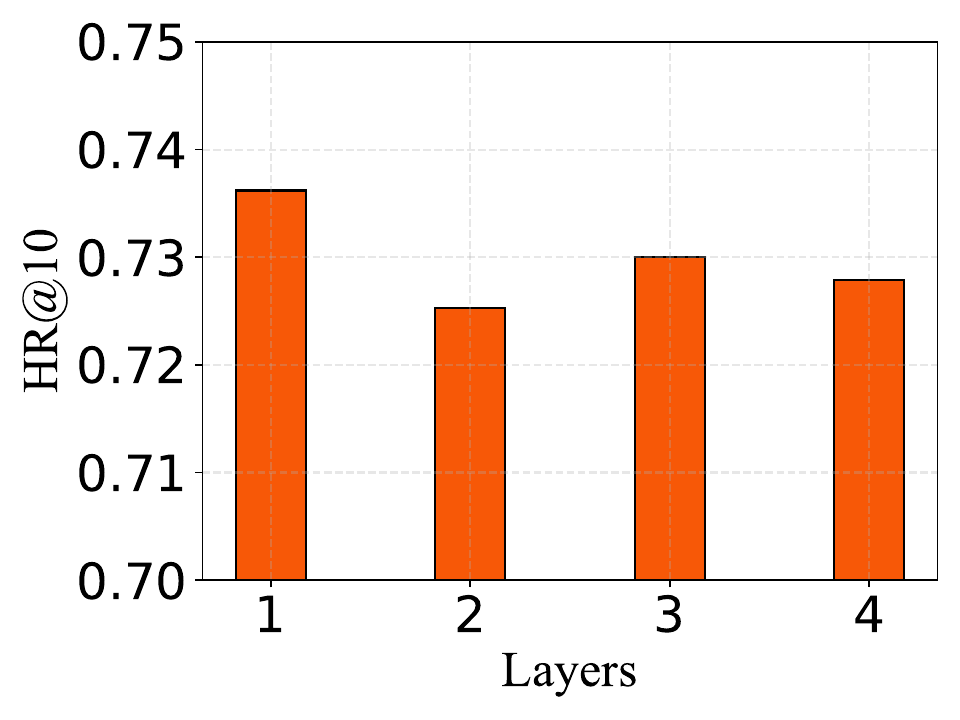}
			% \label{F5a}
	\end{minipage}}
	\subfigure[Performance of NDCG]{
		\begin{minipage}[t]{0.48\linewidth}
			\includegraphics[width=1\linewidth]{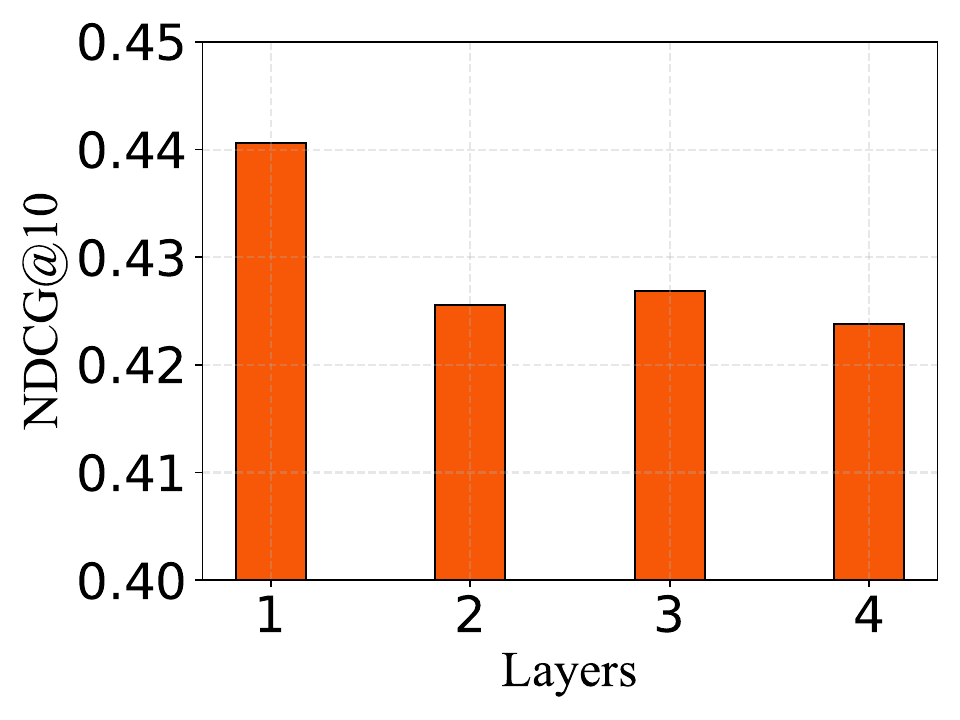}
			% \label{F5b}
	\end{minipage}}%
	\caption{Effect of layers size. We present the HR and NDCG results for both recommendation metrics on the MovieLens-100K dataset.}
	\label{layer}
	% \vspace{-2mm}
\end{figure}

\textbf{\textit{Size of layers.}}
Our proposed method enhances recommendation performance by optimizing item embeddings within a user graph aggregation module. We employ a lightweight Graph Convolutional Network (GCN) to aggregate neighbor information and systematically evaluate the impact of varying GCN layer configurations on recommendation accuracy. Specifically, we experiment with layer counts of 1, 2, 3, and 4, with results summarized in Figure~\ref{layer}. Our findings reveal that a single-layer GCN, which aggregates item embeddings from the most relevant users, consistently outperforms models with additional layers. This suggests that one-hop neighbor information is most critical for effective item recommendation, as deeper layers do not yield further performance gains. These results underscore the importance of prioritizing highly relevant user information in the aggregation process.

We conducted additional hyper-parameter analysis to evaluate the impact of the public users ratio and the personalization level (\(\alpha\)) on recommendation performance. The detailed experimental results are presented in \textbf{Appendix} \ref{other Hyper-parameters}.

\subsection{Recommendation Performance for Public and Private Users \textbf{\textit{(Q4)}}} 
To rigorously evaluate the performance of GFed-PP in realistic mixed-privacy settings, we systematically investigate its effectiveness by varying the proportion of public users. Specifically, we assess the model's performance for both public and private users across a range of public user ratios, with comprehensive results detailed in Table \ref{tab:public_private_ratio} in \textbf{Appendix} \ref{public_private_ratio}.

Our findings indicate that public users derive greater benefits from participating in global model updates. Nevertheless, GFed-PP consistently delivers robust and competitive performance for private users, even in scenarios where public users are scarce. These results underscore the model's resilience and efficacy across heterogeneous privacy configurations.

\subsection{Privacy Protection with Local Differential Privacy \textbf{\textit{(Q5)}}}
To further enhance user privacy, we incorporate Local Differential Privacy (LDP) \cite{ldp} into GFed-PP. We vary the noise strength parameter $\delta \in \{0, 0.1, 0.2, 0.3, 0.4, 0.5\}$ and report the corresponding performance results in Table~\ref{noise}. As expected, increasing the noise level leads to a gradual decline in recommendation accuracy. However, the performance degradation remains limited when the noise is moderate. For example, at $\delta = 0.4$, the model retains most of its predictive power while offering a stronger privacy guarantee. This demonstrates that GFed-PP can effectively balance recommendation quality and privacy by tuning the noise strength.

\begin{table}[htbp]
    \centering

    \caption{Results of using local differential privacy technique into GFed-PP with various noise strength $\delta$.}
    \label{noise}
    \scalebox{0.75}{
    \begin{tabular}{llcccccc}
        \toprule
        Datasets & Noise strength
        & $\delta=0$ & $\delta=0.1$ & $\delta=0.2$ & $\delta=0.3$ & $\delta=0.4$ & $\delta=0.5$ \\
        \midrule
        \multirow{2}{*}{ML-100K} & HR@10 & \textbf{73.62} & 73.11 & 72.83 & 72.51 & 72.24 & 71.83 \\
        & NDCG@10 & \textbf{44.06} & 43.29 & 43.10 & 42.96 & 42.40 & 42.32 \\
        \hline
        \multirow{2}{*}{ML-1M} & HR@10 & \textbf{73.53} & 73.42 & 72.45 & 73.14 & 72.67 & 72.58 \\
        & NDCG@10 & \textbf{44.41} & 43.70 & 43.70 & 44.10 & 43.22 & 43.29 \\
        \hline
        \multirow{2}{*}{Lastfm-2K} & HR@10 & \textbf{90.13} & 89.66 & 89.56 & 89.53 & 88.75 & 88.44 \\
        & NDCG@10 & \textbf{82.79} & 82.00 & 81.95 & 81.96 & 80.78 & 80.10 \\
        \hline
        \multirow{2}{*}{HetRec2011} & HR@10 & \textbf{73.28} & 72.34 & 72.27 & 72.02 & 72.03 & 71.94 \\
        & NDCG@10 & \textbf{46.37} & 45.21 & 45.81 & 45.27 & 45.07 & 45.46 \\
        \hline
        \multirow{2}{*}{Amazon} & HR@10 & \textbf{82.83} & 81.79 & 81.88 & 81.71 & 81.76 & 81.65 \\
        & NDCG@10 & \textbf{73.61} & 71.60 & 72.84 & 72.07 & 72.70 & 71.85 \\
        \bottomrule
    \end{tabular}}
    
\end{table}

\section{Related Works}
\subsection{Centralized Recommendation}
Recommendation systems \cite{2013recommender,recommendation2022} have proven to be effective tools for providing personalized content recommendations (such as movies and items) to users. Their core functionality involves using users' historical behavior data (including clicks and purchase records) to train a global recommendation model. In the field of recommendation system research \cite{MF,ncf2017,wu2022graph}, most approaches can generally be categorized into matrix factorization (MF), deep learning, and graph neural network (GNN) methods. The core of matrix factorization methods \cite{MF,2018surveymf} is to decompose the user-item interaction matrix into two low-dimensional matrices, which represent the latent features of users and items. Deep learning methods \cite{2016deep,2017deep,deldjoo2025toward}, on the other hand, aim to uncover intricate representations from interaction data using deep neural networks, often revealing nonlinear relationships. In recent years, GNN-based recommendation systems \cite{2020lightgcn,2022graph,yang2024local,zhao2025graph} have made significant progress due to their exceptional ability to capture high-order graph structure information. However, these approaches predominantly rely on centralized data processing, which necessitates the collation of user data for model training, thereby posing significant privacy risks
\subsection{Federated Recommendation Systems}
FedRecs \cite{federated2023,federatedzhang2023comprehensive,li2023federated,zhang2024,2024refer} have rapidly emerged as a promising approach, offering personalized recommendation services within a federated learning framework while protecting user privacy. Several benchmark recommendation methods have been adapted to the federated learning framework, including matrix factorization-based approaches such as FCF \cite{FCF}, FedMF \cite{FedMF}, MetaMF \cite{MetaMF}, and FedRecon \cite{FedRecon}, as well as neural collaborative filtering-based methods like FedNCF \cite{FedNCF}. Zhang et al. \cite{PFedRec} proposed the PFedRec framework, a federated personalized recommendation system that eliminates user embeddings in favor of learning personalized rating functions to capture user preferences. Moreover, Zhang et al. \cite{GPFedRec} introduced the GPFedRec framework, which constructs a user graph based on cosine similarity to enhance the learning of personalized preferences. The aforementioned federated recommendation methods assume a uniform privacy-preserving approach for all users. Yan et al. \cite{federatedhgnn} employ the shared heterogeneous graph on the server to assist in the collaborative training of the recommendation model while ensuring user privacy. Wu et al. \cite{wang2025p4gcn} introduce P4GCN, a vertical federated social recommendation framework that employs privacy-preserving two-party graph convolutional networks to enhance recommendation accuracy.
In contrast to prior work, GFed-PP introduces a personalized privacy protection framework that leverages public user data to improve recommendation accuracy without compromising private user information.

\section{Conclusions}
This paper introduces GFed-PP, a novel federated recommendation framework tailored to diverse user privacy requirements. GFed-PP differentiates between public and private users, leveraging publicly shared user-item interaction data to construct a server-side user relationship graph. To preserve privacy, user embeddings are kept entirely local, while only item embeddings are uploaded to the server for collaborative training. Through graph-based aggregation, GFed-PP jointly learns user-specific and global item representations, enabling effective modeling of both public and private user preferences within a unified framework. Extensive experiments on five benchmark recommendation datasets demonstrate that GFed-PP consistently outperforms state-of-the-art centralized and federated recommendation baselines in terms of recommendation accuracy. These results confirm the practical value of GFed-PP as a scalable, flexible, and generalizable solution for privacy-preserving recommendation in real-world scenarios. GFed-PP provides a principled way to balance personalized privacy protection and recommendation accuracy, which is vital for recommendation systems in sensitive user contexts. Future research directions are discussed in \textbf{Appendix}~\ref{Limitations}.
\bibliographystyle{ACM-Reference-Format}
\bibliography{sample-base}

%%
%% If your work has an appendix, this is the place to put it.
\appendix

\section{Notations \label{aaa}}
To facilitate understanding, we summarize the key notations used for the paper in Table~\ref{notations}. These notations cover user and item sets, model components, graph structures, and personalized representations involved in our framework.

\begin{table}[htbp]
    \centering
    \caption{Notation table.}
    \label{notations}
    \begin{tabular}{lc}
        \hline
        \textbf{Notation} & \textbf{Descriptions} \\ \hline
        ${U}$ & The user set \\
        ${I}$ & The item set \\
        $r_{um}$ & The rating of user $u$ on item $m$ \\
        $\hat{r}_{um}$ & The prediction of score function \\
        $M_\theta$ & The recommendation model parameterized with $\theta$ \\
        $G({U},{E})$ & The user relationship graph \\
        $I_i$ & The set of interacted positive items for $i$-th client \\
        $I_i^-$ & The set of sampled negative items for $i$-th client \\
        $\mathbf{A}$ & The adjacency matrix of user relationship graph \\
        $\mathbf{S}$ & User-specific item embeddings \\
        $q_g$ & The global item embeddings \\
        $\mathbf{Q_u}$ & The item embeddings for all users \\
        $q_{u_i}$ & The user-personalized item embedding for $i$-th user \\
        $N$ & The number of clients (users) \\
        $p_i$ & The user embedding module parameter of $i$-th client \\
        $q_i$ & The item embedding module parameter of $i$-th client \\
        $o_i$ & The score function module parameter of $i$-th client \\
        $s_i$ & The user-specific item embedding of $i$-th client \\
        \hline
    \end{tabular}

\end{table}
\section{Algorithm \label{alg}}

We summarize the training procedure of our proposed GFed-PP in Algorithm~\ref{alg: GFed-PP}. The training process alternates between a centralized server-side update based on the user relationship graph and a decentralized client-side update that preserves user-specific personalization and privacy. At each global round, the server constructs a user graph leveraging public user interaction data, applies a graph-based encoder to refine item representations, and learns both global and personalized item embeddings. Each client then updates its local model using received embeddings and individual preferences. This joint optimization enables GFed-PP to effectively balance recommendation accuracy and privacy preservation between users with diverse data sharing preferences.

\begin{algorithm}
\caption{Graph Federated learning for Personalized Privacy Recommendation}
\label{alg: GFed-PP}
\begin{algorithmic}[1]
\STATE Initialize $\alpha, \eta, l, \{(p^{(1)}_i, q^{(1)}_i, o^{(1)}_i)\}_{i=1}^N$
\STATE Initialize $\{s^{(1)}_i\}_{i=1}^N \gets \{q^{(1)}_i\}_{i=1}^N$

\FOR{$e = 1, 2, \ldots, E$}
    \STATE \underline{\textbf{\textit{Server update by user relationship graph:}}}
    \STATE Build user relationship graph $G$ by interaction matrix with Eq. \eqref{eq: user graph} in server
    \STATE Removal of self-loops for public users and insertion of self-loops for private users within the graph $G$ results in a user relationship adjacency matrix $\mathbf{A}$
    \STATE Normalized user relationship adjacency matrix $\mathbf{A}$ with Eq. \eqref{eq: nor}
    \STATE Learn user-specific item embeddings $\{s^{(e+1)}_i\}_{i=1}^N$ with Eq. \eqref{eq: gcn}
    \STATE Learn global item embedding $q_{g}$ with Eq. \eqref{eq: global}
    \STATE Learn user-personalized item embedding $q_{u_i}$ with Eq. \eqref{eq: personal}
    \STATE \underline{\textbf{\textit{Personalized client update:}}}
    \FOR{each client $i = 1, 2, \ldots, N$ in parallel}
        \FOR{$t = 1$ \textbf{to} $T$}
            \STATE Update $(p^{(e)}_i, q^{(e)}_i, o^{(e)}_i)$ with Eq. \eqref{eq: sgd}
        \ENDFOR
        \STATE $(p^{(e+1)}_i, q^{(e+1)}_i, o^{(e+1)}_i) \gets (p^{(e)}_i, q^{(e)}_i, o^{(e)}_i)$
    \ENDFOR
    
\ENDFOR
\end{algorithmic}
\end{algorithm}
\section{Baselines \label{b1}}
We introduce the details of the baselines as follows.
\begin{itemize}
\item \textbf{Matrix Factorization (MF)}\cite{MF}: MF factorizes the user-item rating matrix into low-dimensional embeddings for users and items, capturing latent factors that drive user preferences and item characteristics for accurate recommendations.
\item \textbf{Neural Collaborative Filtering (NCF)}\cite{ncf2017}: NCF is a pioneering neural recommendation model that uses a multi-layer perceptron (MLP) to learn complex interactions between user and item embeddings, providing the flexibility to capture intricate relationships in the data.
\item \textbf{FedMF}\cite{FedMF}: FedMF adapts matrix factorization to federated learning, where user embeddings are updated locally on individual clients and item embedding gradients are aggregated globally on the central server. This privacy-preserving approach ensures user data remains localized while enabling collaborative learning across distributed clients.
\item \textbf{FedNCF}\cite{FedNCF}: FedNCF extends NCF to federated settings, where users update their embeddings locally and share item embeddings and the score function with the server. This enables global aggregation of item information while preserving user privacy.
\item \textbf{Federated Reconstruction (FedRecon)}\cite{FedRecon}: FedRecon is an advanced federated learning framework for personalized recommendation systems. Unlike methods like FedMF, it retrains user embeddings from scratch in each round, ensuring fresh initialization and more personalized updates.
\item \textbf{Meta Matrix Factorization (MetaMF)}\cite{MetaMF}: MetaMF introduces a distributed matrix factorization framework that uses a meta-network to generate the rating prediction module and private item embeddings. The meta-network enables adaptive learning of model parameters, providing a flexible and scalable approach to matrix factorization in federated settings.
\item \textbf{Federated LightGCN (FedLightGCN)}: We adapt LightGCN \cite{2020lightgcn} to the federated learning framework. Specifically, each participating client independently trains a local LightGCN model using its first-order interaction subgraph.
\item \textbf{Federated Graph Neural Network (FedPerGNN)} \cite{fedgnn}: A graph neural network is deployed on each client, and users on the client can incorporate higher-order user-item information via the graph extension protocol.
\item \textbf{Personalized Federated Recommendation (PFedRec)}\cite{PFedRec}: PFedRec is a personalized federated recommendation framework where the server learns a common item embedding for all clients, and each client then fine-tunes the embedding using local data.
\item \textbf{Graph-Guided Personalization Federated Recommendation (GPFedRec)}\cite{GPFedRec}: GPFedRec constructs a user graph based on the similarity of item embeddings. It fine-tunes personalized item embeddings locally on each client and applies a graph-guided aggregation mechanism for federated optimization, preserving privacy while improving recommendation performance.
\end{itemize}

\section{Implementation Details \label{c1}}
During the model training phase, we randomly select four negative samples for each positive sample from items the user did not interact with, applying this approach consistently across all methods. These models are trained using the Stochastic Gradient Descent (SGD) optimizer, valued for its simplicity and effectiveness in optimizing deep learning models.
We unify the embedding size to 32 for all methods, and other details of the baseline models are adhered to as described in their original papers. We employ a fixed batch size of 256 and search for an appropriate learning rate within the range of $[0.0001, 0.001, 0.01, 0.1]$ based on the performance of the validation set. We set the total number of training epochs for centralized methods or communication rounds for federated methods as $100$ to ensure convergence for all methods. An exception is FedPerGNN, where we follow the official code from the original paper for the experimental setup. The number of communication rounds is $3$, as we observe in our experiments that increasing the number of communication rounds did not yield any performance improvement. We implement our method using the PyTorch framework and conduct all experiments with five repetitions, reporting the average results. We conduct all the experiments on 3090 GPUs.

\section{Additional Hyper-parameters Analysis \textbf{\textit{(Q3)}} \label{other Hyper-parameters}}

\begin{figure}
	\centering
	\subfigure[Performance of HR] {
		\begin{minipage}[t]{0.48\linewidth}
			\centering
			\includegraphics[width=1\linewidth]{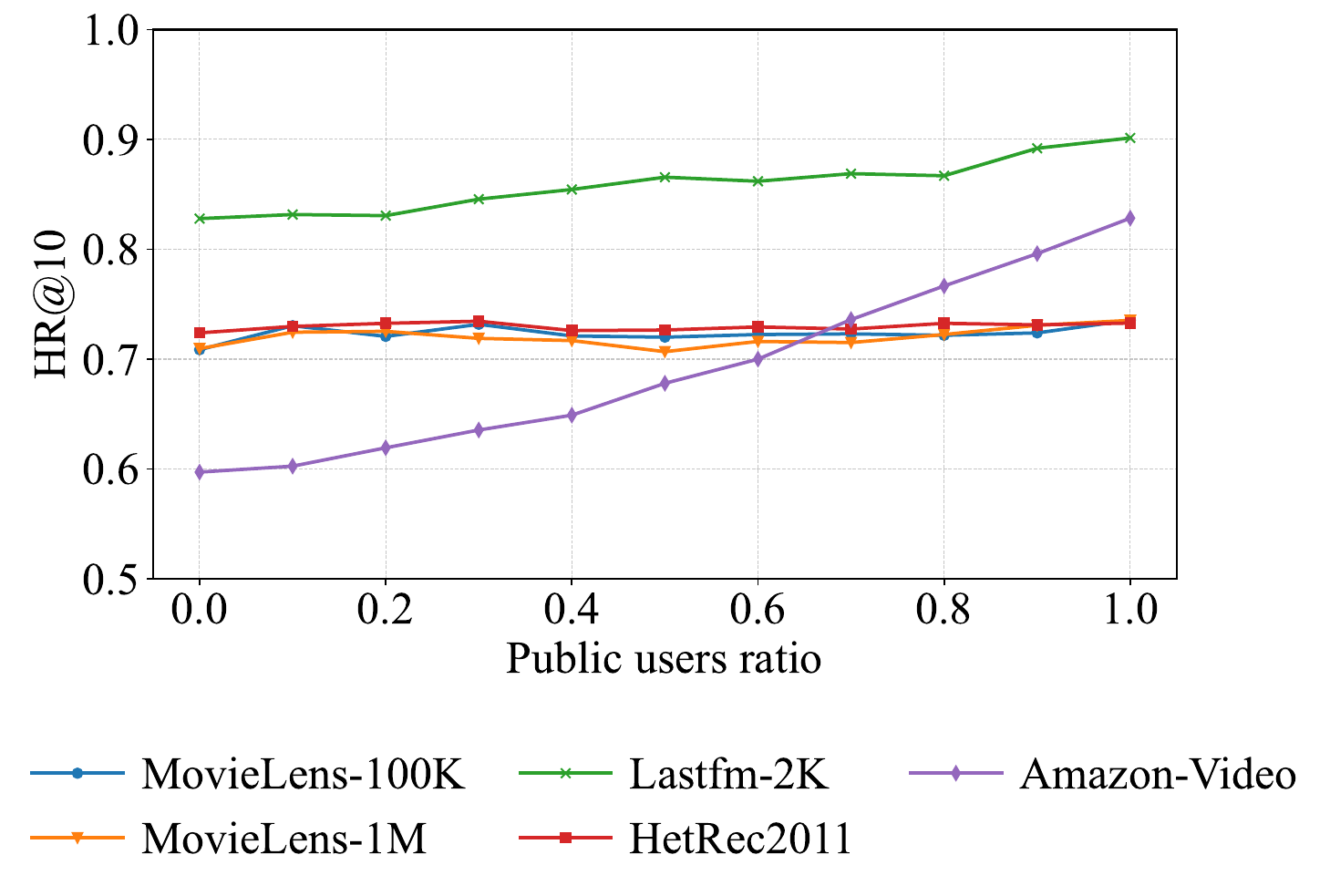}
			% \label{F5a}
	\end{minipage}}
	\subfigure[Performance of NDCG]{
		\begin{minipage}[t]{0.48\linewidth}
			\includegraphics[width=1\linewidth]{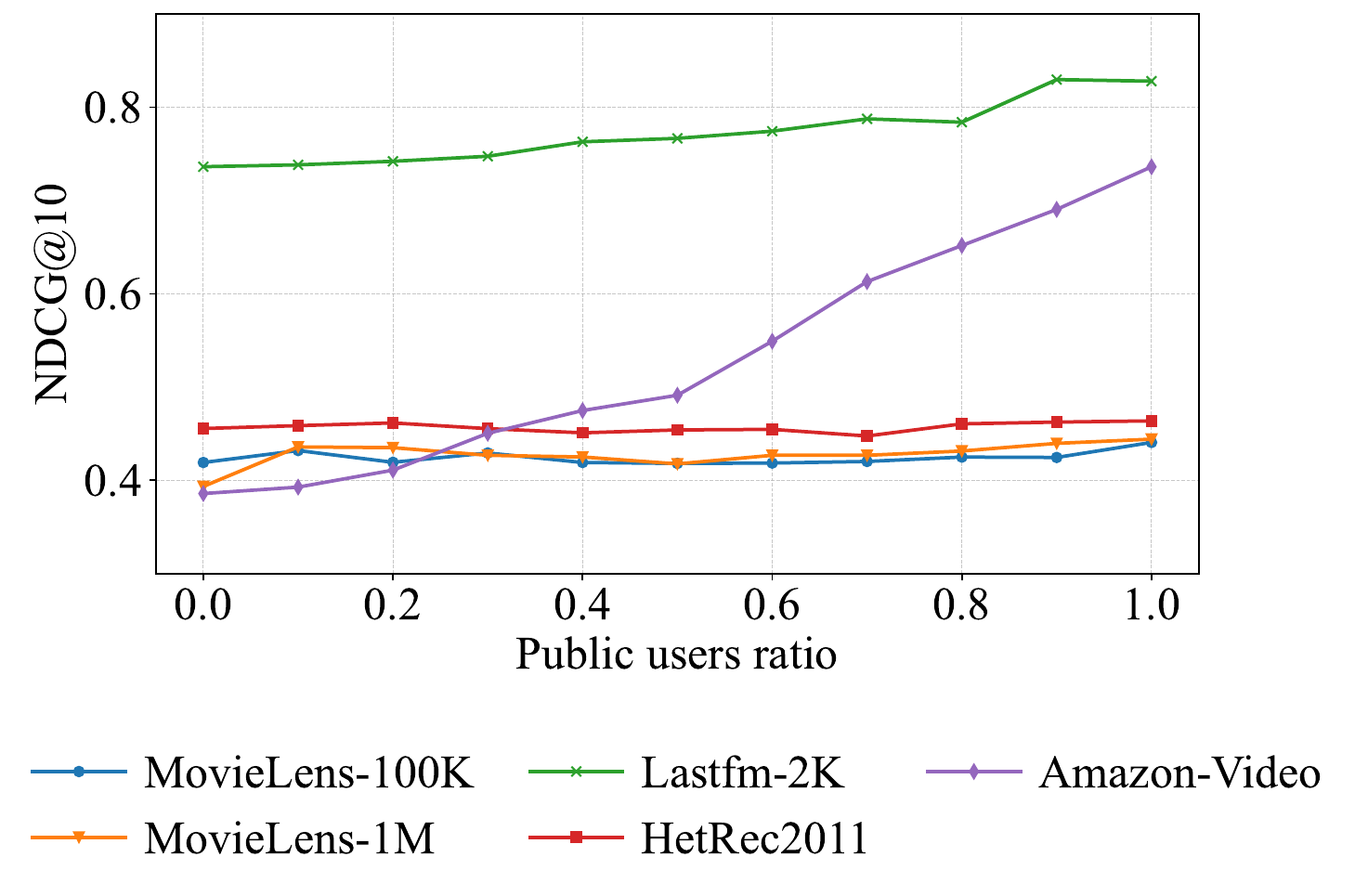}
			% \label{F5b}
	\end{minipage}}%
	\caption{Effect of public users ratio. We present the HR and NDCG results for both recommendation metrics on five datasets.}
	\label{public users ratio}
	% \vspace{-2mm}
\end{figure}

\textbf{\textit{Public users ratio.}}
We investigate how varying the proportion of public users affects model performance. Specifically, we explore the effect of the public users ratio ranging from 0.0 to 1.0 in intervals of 0.1. The results, shown in Figure \ref{public users ratio}, indicate that performance generally improves as the ratio of public users increases. This suggests that when the proportion of public users is small and shared information is limited, the effect of graph aggregation is less significant. As the proportion of public users increases, the performance on the Amazon-Video and Lastfm-2K datasets shows a significant improvement, while the MovieLens-100K, MovieLens-1M, and HetRec2011 datasets exhibit only a modest increase.

\textbf{\textit{The level of personalization ($\alpha$) and public users ratio.}}
\begin{figure}
	\centering
	\subfigure[Performance of HR] {
		\begin{minipage}[t]{0.48\linewidth}
			\centering
			\includegraphics[width=1\linewidth]{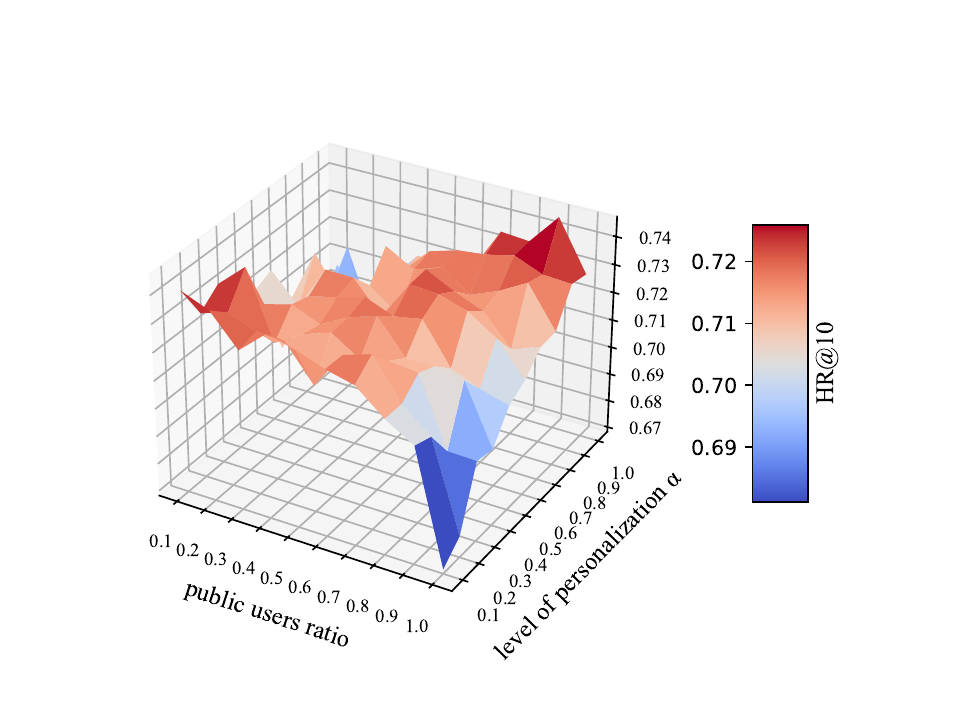}
			% \label{F5a}
	\end{minipage}}
	\subfigure[Performance of NDCG]{
		\begin{minipage}[t]{0.48\linewidth}
			\includegraphics[width=1\linewidth]{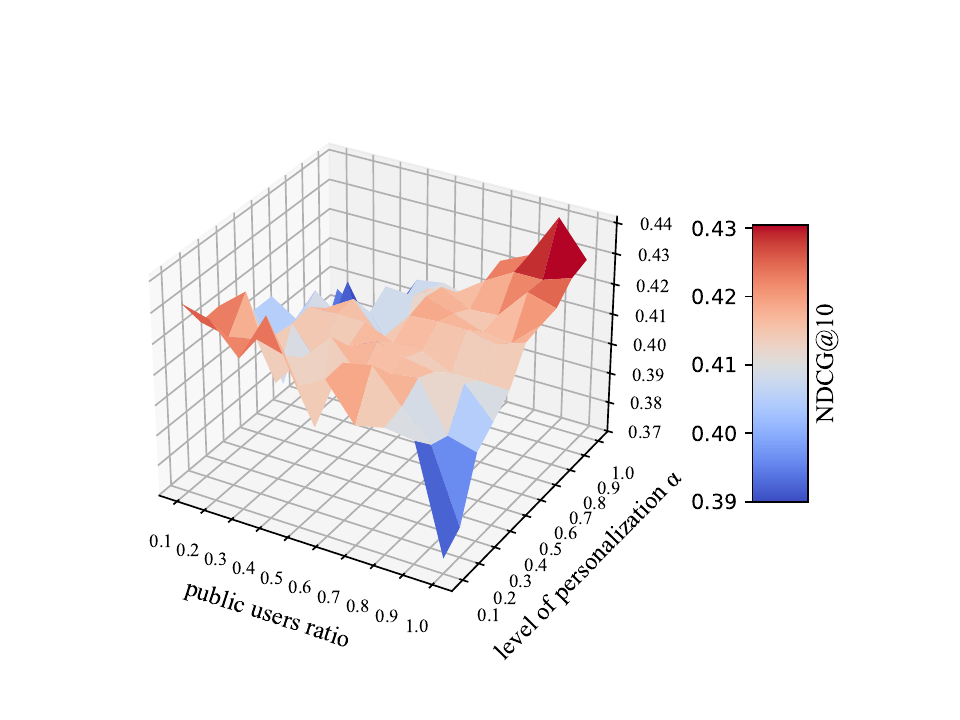}
			% \label{F5b}
	\end{minipage}}%
	\caption{Effect of public users ratio and the level of personalization ($\alpha$). We present the HR and NDCG results for both recommendation metrics on the MovieLens-100K dataset.}
	\label{3d}
	
\end{figure}
We investigate the interaction between the proportion of public users and the level of personalization ($\alpha$), as both factors collectively impact the model performance. We conduct detailed experiments to investigate their interaction. Specifically, we examine the effect of varying public user ratios and $\alpha$ values within the range $[0, 1]$, with intervals of 0.1, on recommendation performance. The results are shown in Figure \ref{3d}, where darker colors indicate better recommendation results. When the proportion of public users is 0.1 and $\alpha$ is also 0.1, the model achieves relatively good performance, indicating that the use of public information benefits personalized recommendations. As the proportion of public users increases, selecting an appropriate value of $\alpha$  allows the model to achieve the best performance. The best recommendation performance occurs when the federated system consists entirely of public users, as shown by the darkest color in the figure.

\section{Recommendation Performance for Public and Private Users \textbf{\textit{(Q4)}} \label{public_private_ratio}}
\begin{table*}[t]
\centering
\caption{Performance of GFed-PP on public and private users under varying public user ratios for Lastfm-2K, HetRec2011, and Amazon-Video datasets.}

\begin{tabular}{lllccccccccc}
\toprule
\textbf{Dataset} & \textbf{User group} & \textbf{Metric} & \textbf{0.1} & \textbf{0.2} & \textbf{0.3} & \textbf{0.4} & \textbf{0.5} & \textbf{0.6} & \textbf{0.7} & \textbf{0.8} & \textbf{0.9} \\
\midrule
\multirow{6}{*}{Lastfm-2K} 
& \multirow{2}{*}{Public users}  & HR@10    & 83.75 & 85.31 & 86.25 & 83.59 & 85.75 & 87.81 & 87.86 & 87.19 & 87.85 \\
&                                  & NDCG@10 & 73.55 & 74.64 & 76.53 & 74.68 & 77.24 & 79.90 & 79.62 & 78.07 & 80.66 \\
& \multirow{2}{*}{Private users} & HR@10    & 83.06 & 83.13 & 83.21 & 83.85 & 83.88 & 82.97 & 83.54 & 84.28 & 84.38 \\
&                                 & NDCG@10 & 74.34 & 72.95 & 73.83 & 73.87 & 74.22 & 72.14 & 72.64 & 72.67 & 73.97 \\
& \multirow{2}{*}{All users}     & HR@10    & 83.13 & 83.56 & 84.21 & 83.75 & 84.81 & 85.88 & 86.56 & 86.38 & 87.50 \\
&                                 & NDCG@10 & 74.26 & 74.96 & 74.90 & 73.89 & 75.82 & 76.31 & 77.12 & 76.61 & 78.34 \\
\midrule
\multirow{6}{*}{HetRec2011} 
& \multirow{2}{*}{Public users}  & HR@10    & 72.24 & 72.24 & 72.30 & 72.38 & 73.30 & 73.72 & 73.36 & 72.78 & 73.01 \\
&                                 & NDCG@10 & 45.86 & 43.81 & 44.47 & 47.57 & 45.83 & 45.13 & 46.74 & 45.57 & 45.92 \\
& \multirow{2}{*}{Private users} & HR@10    & 70.14 & 70.26 & 70.62 & 72.03 & 72.34 & 71.45 & 72.10 & 72.15 & 72.32 \\
&                                 & NDCG@10 & 45.34 & 45.24 & 45.31 & 43.62 & 44.58 & 44.20 & 43.75 & 45.51 & 45.51 \\
& \multirow{2}{*}{All users}     & HR@10    & 72.03 & 72.03 & 72.11 & 72.17 & 72.77 & 71.59 & 72.81 & 72.85 & 72.89 \\
&                                 & NDCG@10 & 45.39 & 44.95 & 45.06 & 45.70 & 44.91 & 45.12 & 45.98 & 45.81 & 46.08 \\
\midrule
\multirow{6}{*}{Amazon-Video} 
& \multirow{2}{*}{Public users}  & HR@10    & 64.31 & 66.19 & 68.08 & 76.08 & 76.09 & 76.09 & 82.12 & 83.21 & 85.20 \\
&                                 & NDCG@10 & 44.36 & 51.34 & 50.21 & 55.23 & 66.94 & 69.64 & 71.12 & 75.57 & 75.86 \\
& \multirow{2}{*}{Private users} & HR@10    & 59.41 & 60.78 & 60.61 & 60.06 & 60.06 & 60.33 & 61.64 & 60.15 & 60.15 \\
&                                 & NDCG@10 & 38.32 & 39.47 & 39.43 & 39.39 & 39.93 & 40.95 & 39.98 & 39.86 & 39.75 \\
& \multirow{2}{*}{All users}     & HR@10    & 59.90 & 61.86 & 63.84 & 66.80 & 69.43 & 72.56 & 75.98 & 81.79 & 81.79 \\
&                                 & NDCG@10 & 38.92 & 42.21 & 43.02 & 49.13 & 53.04 & 57.94 & 62.07 & 66.85 & 72.24 \\
\bottomrule
\end{tabular}

\label{tab:public_private_ratio}
\end{table*}
To simulate the real-world distribution of public and private users, we vary the proportion of public users from $0.1$ to $0.9$ and evaluate the recommendation performance for both user groups. Detailed experimental settings are provided in Table \ref{tab:public_private_ratio}.

The results in Table \ref{tab:public_private_ratio} show that while public users generally benefit from higher recommendation performance due to participation in global model updates, GFed-PP still achieves competitive results for private users across all settings. As the proportion of public users increases, overall performance improves, but the model maintains stable accuracy for private users, demonstrating its robustness and effectiveness in different privacy scenarios.

\section{Future Research Directions \label{Limitations}}

Future research will aim to enhance the realism of the evaluation framework. One direction is to collect datasets with user-annotated privacy preferences or to develop simulation protocols grounded in empirical studies of user behavior. These efforts will improve the validity and generalizability of privacy-aware recommendation models in real-world scenarios.
To address the fairness gap, future work will also explore fairness-aware optimization techniques tailored for federated environments. Potential solutions include personalized loss reweighting, domain adaptation, or calibration mechanisms to ensure balanced performance across users with varying privacy levels.
Finally, incorporating large language models (LLMs) \cite{zhao2023survey} into GFed-PP offers a promising avenue. LLMs can elicit user privacy preferences through natural language interfaces and support adaptive, interpretable recommendations. 
\end{document}